# Automated Real-time Assessment of Intracranial Hemorrhage Detection AI Using an Ensembled Monitoring Model (EMM)

Zhongnan Fang[1,2], Andrew Johnston[1,2], Lina Cheuy[1,2], Hye Sun Na[1,2], Magdalini Paschali[1,2], Camila Gonzalez[1,2], Bonnie A. Armstrong[1,2], Arogya Koirala[1,2], Derrick Laurel[1,3], Andrew Walker Campion[1], Michael Iv[1], Akshay S. Chaudhari[1,2,4] *, David B. Larson[1,2] *

* denotes co-senior authorship

**Affiliations:**

1. Department of Radiology, School of Medicine, Stanford University, Stanford, CA 94304
2. AI Development and Evaluation Laboratory (AIDE), School of Medicine, Stanford University, Stanford, CA 94304
3. 3D and Quantitative Imaging Laboratory (3DQ), School of Medicine, Stanford University, Stanford, CA 94304
4. Department of Biomedical Data Science, School of Medicine, Stanford University, Stanford, CA 94304

## Abstract

Artificial intelligence (AI) tools for radiology are commonly unmonitored once deployed. The lack of real-time case-by-case assessments of AI prediction confidence requires users to independently distinguish between trustworthy and unreliable AI predictions, which increases cognitive burden, reduces productivity, and potentially leads to misdiagnoses. To address these challenges, we introduce Ensembled Monitoring Model (EMM), a framework inspired by clinical consensus practices using multiple expert reviews. Designed specifically for black-box commercial AI products, EMM operates independently without requiring access to internal AI components or intermediate outputs, while still providing robust confidence measurements. Using intracranial hemorrhage detection as our test case on a large, diverse dataset of 2919 studies, we demonstrate that EMM successfully categorizes confidence in the AI-generated prediction, suggesting different actions and helping improve the overall performance of AI tools to ultimately reduce cognitive burden. Importantly, we provide key technical considerations and best practices for successfully translating EMM into clinical settings.

# Introduction

The landscape of healthcare has rapidly evolved in recent years, with an exponential increase in FDA-cleared artificial intelligence (AI) software as medical devices, especially in radiology[1]. Despite the number of AI applications available, the clinical adoption of radiological AI tools has been slow due to safety concerns regarding potential increases in misdiagnosis, which can erode overall trust in AI systems[2]. Such inaccurate predictions force meticulous verification of each AI result, ultimately adding to the user's cognitive workload rather than fulfilling AI's promise to enhance clinical efficiency[3]. This mismatch between expected and actual performance of AI tools indicates a clear need for real-time monitoring to inform physicians on a case-by-case basis about how confident they can be in each prediction. Such real-time monitoring alongside physician's image interpretation also goes hand-in-hand with the latest guidance issued by the FDA focusing on total life-cycle management of AI tools, rather than the status-quo of pre-deployment validation[4]. However, there are currently limited guidelines or best practices for real-time monitoring to communicate reduced model confidence or uncertainty in an AI model's prediction.

Current monitoring of radiological AI devices is performed retrospectively based on concordance between AI model outputs and manual labels, which require laborious radiologist-led annotation[5]. Due to the resource-intensive nature of generating these labels, the vast majority of retrospective evaluations are limited to small data subsets, providing only a partial view of real-world performance[6]. While recent advances in large language models (LLMs) have shown promise in the analysis of clinical reports[7–9], including automated extraction of diagnosis labels from radiology reports[10–12], this solution remains retrospective. Moreover, with report-based monitoring, regardless of the extraction technique, the "quality control" mechanism for algorithm performance remains a manual task. An automated quality monitoring solution may help decrease user cognitive burden and provide additional objective information regarding the performance of the AI model, including performance drift.

In response to these limitations, various real-time monitoring techniques have been proposed, including methods that can directly predict confidence/uncertainty using the same training dataset used to develop the AI model being monitored[13–18]. Common classes of confidence/uncertainty estimation rely on methods such as SoftMax probability calibration[19–21], Bayesian neural networks[22–25], and Monte Carlo dropout[26,27]. Deep ensemble approaches have also emerged to evaluate prediction reliability by utilizing groups of models derived from the same parent model with varying augmentations[28–32]. However, these methods require access to either the training dataset, model weights, or intermediate outputs, which is not practical when monitoring commercially available models. Since all FDA-cleared radiological AI models that are deployed clinically are black-box in nature, there currently exist no techniques to monitor such models in production in real time.

Thus, there remains a critical need for a real-time monitoring system to automatically characterize confidence at the point-of-care (i.e., when the radiologists review the images and the black-box AI prediction in question). To address this need, we developed the Ensembled Monitoring Model (EMM) approach, which is inspired by clinical consensus practices, where individual opinions are validated through multiple expert reviews. Our EMM framework enables prospective real-time case-by-case monitoring, without requiring ground-truth labels or access to internal AI model components, making it deployable for black-box systems. Here, we demonstrate the effectiveness of the EMM approach in characterizing confidence of intracranial hemorrhage (ICH) detection AI systems (one FDA-cleared and one open-source) operating on head computed tomography (CT) imaging. In this clinically significant application requiring high reliability, we show how EMM can monitor AI model performance in real time and inform subsequent actions in cases flagged for decreased accuracy. The complementary use of a primary AI model with EMM can improve accuracy and user trust in the AI model, while potentially reducing the cognitive burden of interpreting ambiguous cases. We further investigate and provide key considerations for translating and implementing the EMM approach across different clinical scenarios.

## Results

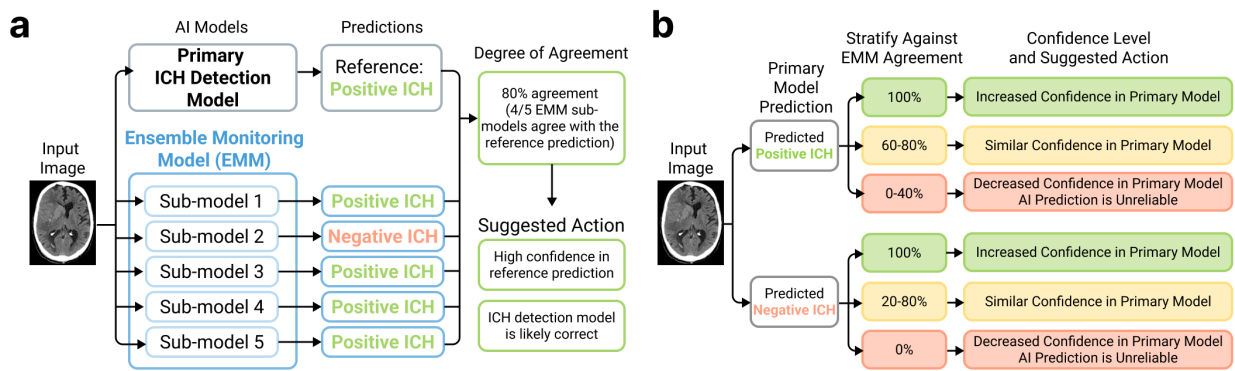

*Figure 1. Overview of Ensembled Monitoring Model (EMM) and example of how to stratify EMM agreement into suggested actions. a.* Each sub-model within the EMM is trained to perform the same task as the primary ICH detection model. The independent sub-model outputs are then used to compute the level of agreement between the ICH detection model and EMM, helping quantify confidence in the reference prediction and suggesting an appropriate subsequent action. *b.* An example EMM use case is stratifying cases into categories of increased, similar, or decreased confidence in the primary model predictions after computing the level of EMM agreement. Through careful selection of the stratification thresholds based on the primary model's performance at different EMM agreement levels, this categorization enables radiologists to make different decisions based on the confidence level derived from the EMM agreement levels.

*Ensembled Monitoring Model (EMM) Overview*

Emulating how clinicians achieve group consensus through a group of experts, the EMM framework was developed to estimate consensus among a group of models. Here, we refer to the model being monitored as the "*primary model*". In this study, the EMM comprised five sub-models with diverse architectures trained for the identical task of detecting the presence of ICH (**Figure 1a**). Each sub-model within the EMM independently processed the same input to generate its own binary prediction (e.g. the input image is positive or negative for ICH), in parallel to the primary ICH detection model. Each of the five EMM outputs were compared to the primary model's output to quantify the agreement between each pair of predictions, from 0% to 100% agreement (meaning that none or all five EMM sub-models agreed with the primary output). This level of agreement can translate into confidence in the primary output.

The level of EMM agreement with the primary ICH detection model and resulting degree of confidence also enables radiologists to make different decisions on a case-by-case basis. EMM's agreement level with the primary model can stratify the primary predictions into three groups to represent cases in which the radiologist can have increased, similar, or decreased confidence after seeing EMM's agreement with the primary model (**Figure 1b**). This stratification allows radiologists to adjust their actions accordingly for each image read. For example, the primary model's prediction might not be used in cases with decreased confidence, and these cases should be reviewed following a radiologist's conventional image interpretation protocol. As we show in the following sections, such optimization may potentially improve radiologist efficiency and reduce cognitive load.

*EMM agreement levels are associated with different features*

To identify the features most commonly found in images with high EMM agreement, we conducted a visual examination of all 2,919 analyzed CT studies (50.1% male, age range: 2 months-104.6 years). Two primary ICH detection models were evaluated: an FDA-cleared AI model, and an open-source AI model that placed second-place[33] in the RSNA 2019 ICH challenge[34]. The two primary AI models complemented each other for evaluation, with the FDA-cleared AI model showing lower sensitivity but higher specificity, precision, and overall accuracy than the open-source model (**sFigure 1**).

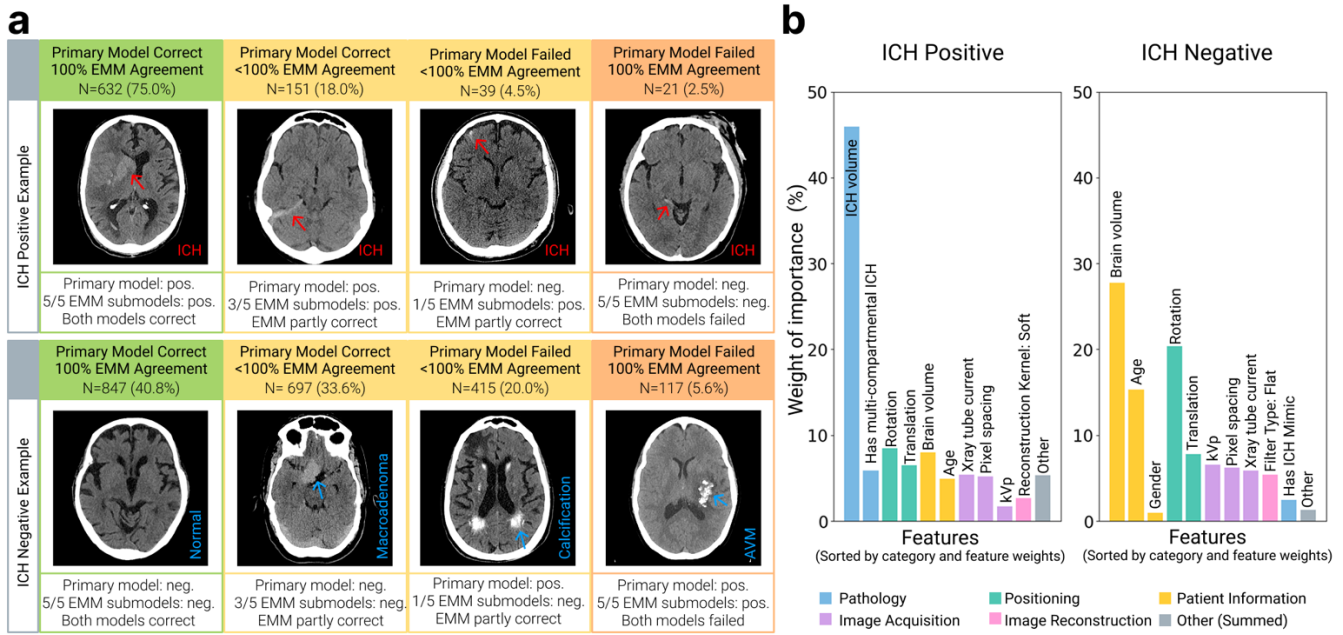

*Figure 2. EMM agreement associated with different features. **a.** Example cases for which EMM showed different levels of agreement with the FDA-cleared primary ICH detection model. Cases with full EMM agreement typically showed clear presence or absence of ICH, while cases with partial agreement often displayed subtle ICH or features mimicking hemorrhage. **b.** Quantitative analysis on the importance of features affecting EMM agreement with the FDA-cleared primary model in ICH-positive and ICH-negative cases. The normalized weight of importance for all features sums to 100%.*

Results of the FDA-cleared primary model are shown in **Figure 2a**, and the results of the open-source model are shown in **sFigure 2**. The FDA-cleared primary model and EMM demonstrated 100% agreement and correct classifications in 1,479 cases (51%, ICH positive 632, negative 847), primarily in cases with obvious hemorrhage or a clearly normal brain anatomy. EMM showed partial agreement with the FDA-cleared model in 848 cases (29%, ICH positive 151, negative 697) when the FDA-cleared model was correct. And EMM also showed partial agreement in 454 cases (16%, ICH positive 39, negative 415) when the FDA-cleared model was incorrect. Visual examination revealed that the cases with partial agreement typically presented with subtle ICH or contained imaging features that mimicked hemorrhage (e.g. hyperdensity, such as calcification or tumor). These cases of partial agreement provide an opportunity for further radiologist review. Finally, in 138 cases (4%, ICH positive 21, negative 117), EMM demonstrated 100% agreement with the FDA-cleared model, but EMM failed to detect that the FDA-cleared model's prediction was wrong. These cases predominantly involved either extremely subtle hemorrhages or CT features that strongly mimicked hemorrhage patterns, confusing both the FDA-cleared model and EMM.

We then quantitatively examined which features affected EMM agreement using Shapley analysis[35]. This analysis was performed on a data subset (N=281) with a comprehensive set of features manually labeled by radiologists spanning multiple categories, including pathology, patient positioning, patient information, image acquisition and reconstruction parameters. In ICH-positive cases, hemorrhage volume emerged as the dominant feature for high EMM agreement, with larger volumes strongly corresponding to higher agreement (**Figure 2b**), as seen in our visual analysis. For ICH-negative cases, the predictive features for EMM agreement were more balanced. The top predictors for EMM agreement were brain volume, patient age, and image rotation. Some directional relationships between feature values and EMM agreement were also identified (**sFigure2**). For ICH-positive cases, high hemorrhage volume and multi-compartmental hemorrhages resulted in higher EMM agreement. In ICH-negative cases, the presence of features that mimicked hemorrhages led to lower EMM agreement.

*EMM Enables Confidence-based Image Review Optimization*

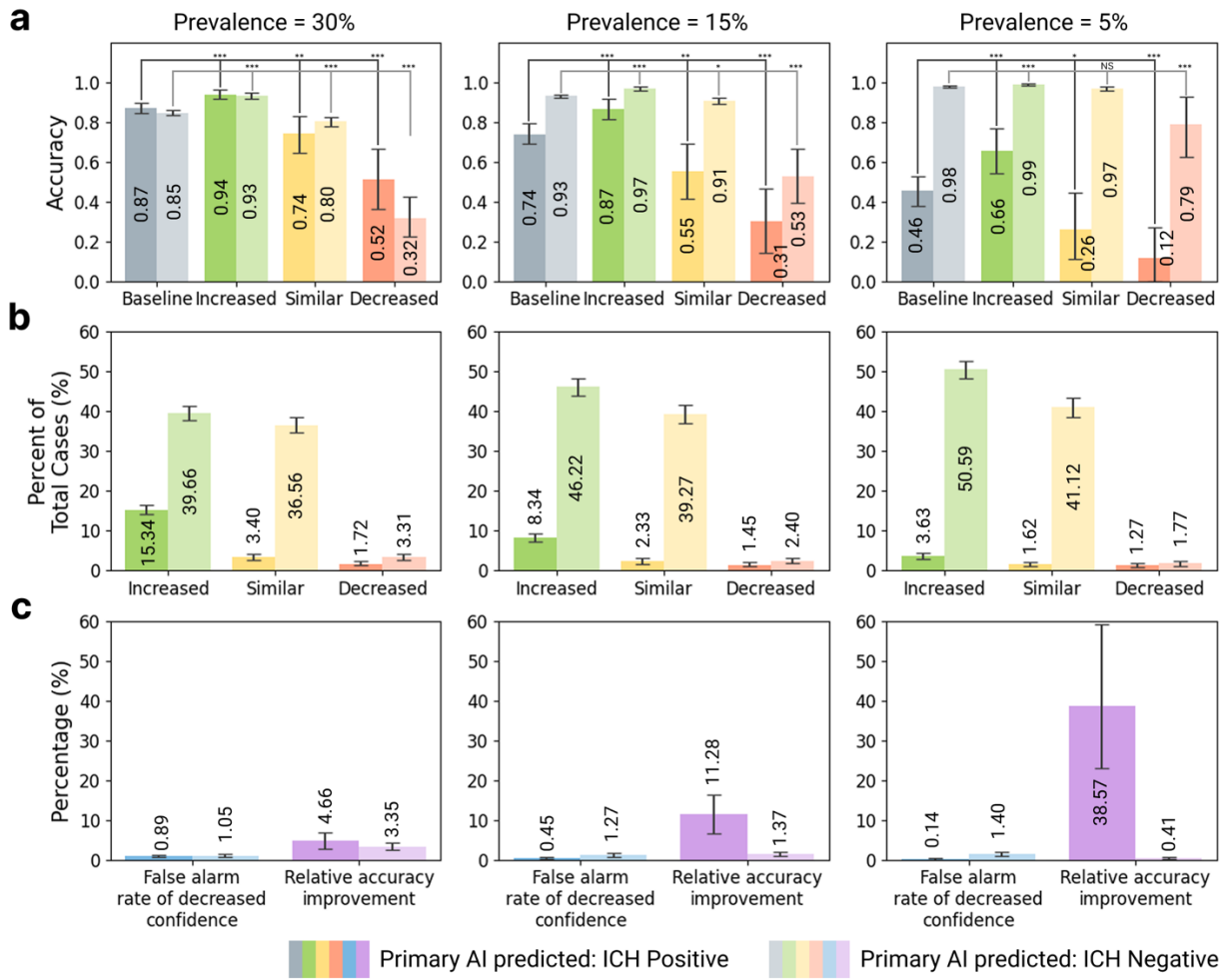

*Figure 3. EMM stratifies cases into different accuracy groups for the FDA-cleared model and enables customized clinical decision making. a. Cases stratified by EMM agreement levels*

*demonstrated increased (green), similar (yellow), or decreased (red) accuracies compared to the baseline accuracy of the primary model without EMM (gray). **b.** Distribution (%) of cases classified as increased (green), similar (yellow), or decreased (red) confidence based on the EMM agreement thresholds. **c.** For the cases in which EMM indicated decreased confidence, a more detailed radiologist review is called for. Cases flagged for decreased confidence, but for which the primary model's prediction was correct, were defined as false alarms. Substantial relative gains over baseline accuracy using only the primary model were observed across all prevalence levels for ICH-positive primary model predictions, outweighing the burden of false alarms. For ICH-negative primary model predictions, however, this favorable balance between relative accuracy gains and false alarm burden was only observed at 30% prevalence.*

Following the EMM thresholds and suggested actions outlined in **Figure 1b**, we established stratification thresholds based on expert radiologist feedback for the FDA-cleared ICH detection model based on its accuracy at different EMM agreement levels, shown in **sFigure 3**. For ICH-positive primary model predictions, we set thresholds of 100% EMM agreement for increased confidence in the primary model, 60% or 80% EMM agreement for similar confidence in the primary model, and 0%, 20% or 40% EMM agreement for decreased confidence in the primary model. For ICH-negative primary model predictions, thresholds were set at 100% EMM agreement for increased confidence, 20%, 40%, 60% or 80% EMM agreement for similar confidence, and 0% EMM agreement for decreased confidence. We then evaluated the overall accuracy of the primary model together with EMM for cases classified as increased, similar, and decreased confidence (**Figure 3a**). This evaluation was also performed across three different prevalences of 30%, 15%, and 5%, which are close to the prevalences observed at our institution across emergency, in-patient, and out-patient settings. As expected, overall accuracy was highest for cases in which the primary model and EMM showed high agreement, and overall accuracy was lowest when EMM showed a lower agreement level with the primary model. This was observed for both ICH-positive and ICH-negative primary predictions and across all prevalence levels. Of the cases analyzed, most of the cases were classified as increased confidence based on EMM thresholds, followed by similar confidence, and lastly decreased confidence (**Figure 3b**).

To assess the practical value of the EMM suggested actions, we analyzed the relative gains of the model compared to the cognitive load and loss of trust associated with incorrect classifications. Among the cases flagged for decreased confidence, those for which the primary model's prediction remained correct despite low EMM agreement (and thus the decreased confidence classification) were considered false alarms. As shown in **Figure 3c**, the potential for radiologists to be alerted to a possible incorrect primary model output and correct these cases substantially improved relative accuracy compared to the percentage of false alarms across all prevalence levels (relative accuracy improvements of 4.7%, 11%, and 38% versus false-alarm rates of 0.89%, 0.45%, and 0.14% at prevalence levels of 30%, 15%, and 5%, respectively). However, this net benefit was only observed for cases with ICH-negative primary model predictions at 30% prevalence (false-alarm rate of 1.1% versus relative accuracy improvement of 3.4%). At lower prevalence levels (15% and 5%), the already-high baseline accuracy of the primary model for ICH-negative cases (i.e. accuracy=0.93 and 0.98, respectively) meant that the burden of false alarms (1.3% and 1.4%) did not exceed the relative accuracy gains (1.4% and 0.41%). Similar results were also observed for the open-source ICH detection AI across all prevalences (**sFigure 4**).

*Sub-model and Data Size Considerations when Training EMM*

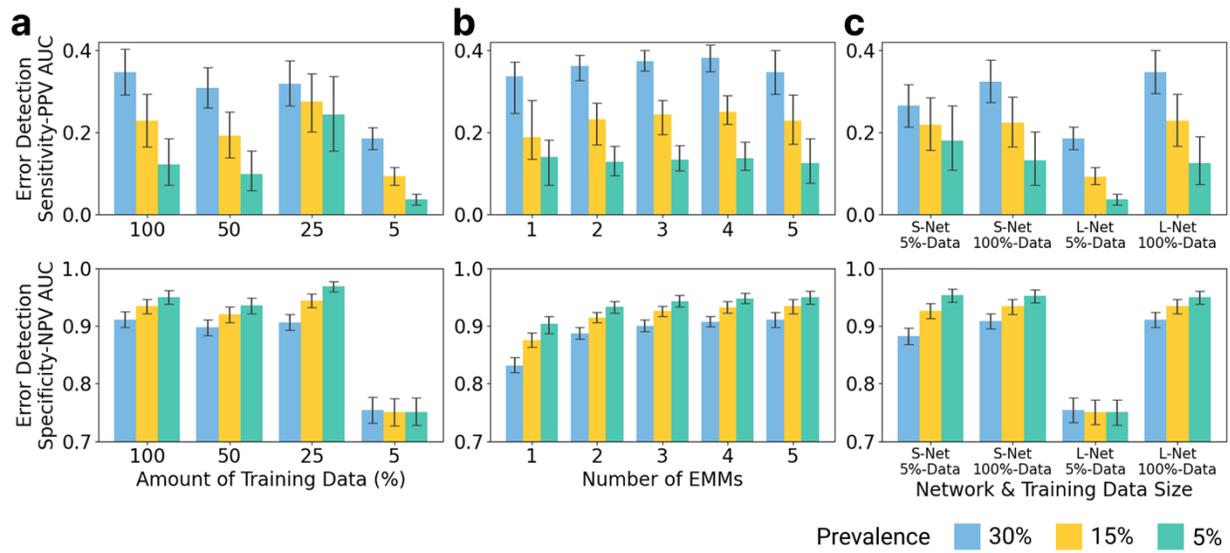

*Figure 4. EMM performance for detecting errors made by the commercial primary model generally improved with increasing (a) training data volume, (b) number of sub-models, and (c) sub-model sizes.* Error detection sensitivity-PPV area under curve (ED-SPAUC) and specificity-NPV area under curve (ED-SNAUC) were measured across prevalences; higher values are desirable for both. S-Net represents small sub-model networks for EMM, L-Net represents large sub-model networks for EMM. Similar results for the open-source ICH model are shown in **sFigure4**.

To enable broader application and adoptability, we conducted a comprehensive analysis of how three key factors affect EMM performance: i) amount of training data used: 100% of the dataset (n=18,370), 50% of the dataset (n=9,185), 25% of the dataset (n=4,592), and 5% of the dataset (n=918), ii) number of EMM sub-models (1-5), and iii) EMM sub-model size in relation to training data volume. EMM performance was measured by its ability to detect errors made by the primary model using error detection sensitivity-PPV area under curve (ED-SPAUC) and specificity-NPV area under curve (ED-SNAUC) across prevalences, as these metrics consider the overall error detection performance regardless of the agreement level threshold applied.

**Training Data**: As illustrated in **Figure 4a**, EMM's ED-SPAUC for the FDA-cleared primary model generally decreased as the training data was reduced from 100% to 5% of the original dataset across all three prevalences. This suggested that EMM generally improves with increased training data volume, though the benefits begin to saturate after approximately 10,000 studies. This trend was also observed in the error detection SNAUC and the open-source primary model (**sFigure 5a**).

**Number of Ensemble Sub-models**: As shown in **Figure 4b**, EMM's ED-SPAUC increased as the number sub-models increased from 1 to 4, before generally stabilizing at 5 across all three prevalence levels. Conversely, ED-SNAUC consistently improved as the number of models increased from 1 to 5, across all prevalences. Similarly, both error detection metrics showed consistent improvement as the number of networks increased for the open-source primary AI (**sFigure 5b**). These results suggest that EMM performance generally increases with more sub-models, with 4 or 5 sub-models serving as an effective starting point for future applications.

**Ensemble Sub-model Size and Training Data**: We next investigated how combinations of EMM sub-model size and training data volume affected EMM's performance in monitoring the primary model. We examined four scenarios: i) ensemble of small networks trained with 5% of the dataset (S-Net 5%-Data), ii) ensemble of small networks trained with 100% of the dataset (S-Net 100%-Data), iii) ensemble of large networks trained with 5% of the dataset (L-Net 5%-Data), and iv) ensemble of large networks trained with 100% of the dataset (L-Net 100%-Data). As shown in **Figure 4c**, EMM generally achieved the best ED-SPAUC and ED-SNAUC with large networks and 100% of the training data and worst performance with large network and 5% of training data, suggesting that a larger training dataset could benefit EMM performance. In a 5% prevalence setting, an ensemble of small networks and 5% of the training data achieved the highest ED-SPAUC and ED-SNAUC values. Similar findings were also observed for the open-source primary AI (**sFigure 5c**).

Taken together, these results provide insights into how the EMM approach can be developed and tailored for various real-time monitoring applications.

## Discussion

As interest in AI for healthcare rapidly grows, monitoring medical AI systems has become increasingly urgent to ensure AI's trustworthiness, safety, and effectiveness. Recently, the FDA has proposed a lifecycle management approach for medical AI devices, requiring evaluation not only during FDA approval and pre-deployment phases, but also continuous monitoring after real-world implementation, and critical risk assessments for individual cases across AI systems[4]. However, the black-box nature of FDA-cleared commercial AI systems creates significant challenges for real-time case-specific monitoring. In this paper, we introduce EMM, a framework that monitors black-box clinical AI systems in real-time without requiring manual labels or access to the primary model's internal components. Using an ensemble of independently trained sub-models that mirror the primary task, our framework measures confidence in AI predictions through agreement levels between the EMM sub-models and the primary model on a case-by-case basis.

These agreement levels can then be used to stratify cases by confidence in the primary model's prediction and suggest a subsequent action. This makes EMM a valuable tool that fills the critical gap in real-time, case-by-case monitoring for FDA-cleared black-box AI systems that would otherwise remain unmonitored.

Our approach enables quantification of confidence through EMM agreement levels with the primary model's predictions. By applying appropriate thresholds to the level of agreement between the EMM and primary model, radiologists can differentiate between which predictions they can be more or less confident in, therefore optimizing their attention allocation and cognitive load. For example, with EMM, radiologists could feel more confident in over half of all cases interpreted. Notably, EMM also reliably identified cases of low confidence, allowing for focused review of these cases and greatly improving overall ICH detection accuracy. In our testing, EMM only failed alongside the primary model in a small percentage of cases (4%). These cases of both EMM and primary model being incorrect represent those with small ICH volumes or ICH-mimicking features.

The stratification of confidence levels based on the EMM agreement levels also enables radiologists to make tailored decisions for each case. The thresholds for defining the three accuracy groups in this study were established based on expert radiologist assessment and the primary model's performance at different EMM agreement levels (**sFigure3**), with separate analyses for ICH-positive and ICH-negative primary model predictions. The thresholds to indicate increased and decreased confidence were specifically designated so that the overall ICH detection accuracy would be significantly higher or lower, respectively, than that with only the primary model (baseline). However, suboptimal EMM agreement thresholds (resulting in too many cases categorized as decreased confidence) can create an unfavorable trade-off where the burden of further reviewing false alarms, and the associated loss in trust in the EMM, outweighs the relative gains in accuracy. This inefficiency particularly impacts low-prevalence settings, where radiologists may waste valuable time reviewing cases that the primary model had already classified correctly **(sFigure 6)**. This illustrates that although the overall EMM framework can be applied to broad applications, the agreement levels and thresholds may need case-specific definitions depending on the prevalence level.

Varying the technical parameters of EMM also revealed insight into the best practices for applying EMM to other clinical use cases. Our ablation study revealed that expectedly, larger datasets, a larger number of sub-models, and larger sub-models generally improve the EMM's capability to detect errors in the primary AI model. We also observed that at 5% prevalence, large sub-models trained with 25% of the data or small model trained with 5% of the data achieved optimal performance. This behavior can be explained by the relationship between model complexity and data volume. Specifically, large sub-models trained on the full dataset (with 41% prevalence) likely became too calibrated/overfitted to that specific prevalence distribution, causing suboptimal performance when testing on data with significantly different prevalence (5%)[36]. Using large sub-model training with 25% of data or using small sub-model training with 5% of data may help the EMM balance the bias-variance tradeoffs by learning meaningful patterns for generalizability, while not overfitting to the training prevalence. The differences observed in optimal dataset size and sub-model size across different prevalence levels can help inform the best technical parameters to start developing an EMM for a different use case, promoting greater adoptability across diseases.

Beyond using EMM to improve case-by-case primary model performance, as shown in this study, another potential application of EMM can be monitoring longitudinal changes in primary AI performance. As the EMM agreement levels are tracked over time, perturbations in the expected ranges can be identified over daily, weekly, or monthly periods. For example, any significant drifts in EMM agreement level distribution may signal changes in primary model performance due to shifts in patient demographics, image acquisition parameters, or clinical workflows. In this manner, the EMM approach can provide another dimension into the current radiology statistical process/quality control pipelines for continuous background monitoring[37,38], in addition to reporting concordance.

While the EMM approach demonstrates several advantages in case-by-case AI monitoring, some limitations persist. Although the EMM does not require labels to perform its monitoring task, a key constraint is the need for labeled use-case-specific datasets when training the EMM for each clinical application, which could potentially limit broader adoption across diverse clinical institutions with different computing resources. However, with the recent maturity of LLMs and self-supervised model training techniques, this limitation may be largely overcome. For example, labels can now be automatically extracted from existing radiology reports using LLMs [10–12]. Self-supervised training[39,40] also enables large foundation model training without manual annotation[41,42], and only a small amount of labeled data would be required to further fine-tune the model for each use case. These recent developments enable periodic updates to EMM, allowing it to adapt to changes in patient populations, scanners, and imaging protocols, thereby maintaining consistent and robust performance over time. Another limitation is EMM's susceptibility to similar failure patterns as the primary model being monitored, such as in cases involving small, low-contrast hemorrhages or ICH-mimicking pathologies in this study. Of particular concern are instances where EMM fails simultaneously with the primary model while indicating complete consensus, as this could instill false confidence in clinicians and potentially increase misdiagnosis risk. This risk might be mitigated by training EMM in the future using synthetic datasets[43] with artificially generated difficult cases, such as those with less obvious hemorrhages, with ICH mimicking features, representing diverse patient population, or with various artifacts. As AI technology rapidly develops, many of the limitations currently facing EMM may be quickly overcome, presenting greater opportunities to not only improve EMM performance but also the resources required to implement the EMM approach itself.

In conclusion, our EMM framework represents a significant advancement in black-box clinical AI monitoring, enabling case-by-case confidence estimation without requiring access to primary model parameters or intermediate outputs. By leveraging ensemble agreement levels, EMM provides actionable insights, potentially enhancing diagnostic confidence while reducing cognitive burden. As AI continues to integrate into clinical workflows, approaches like EMM that provide transparent confidence measures will be essential for maintaining trust, ensuring quality, and ultimately improving patient outcomes in resource-constrained environments.

## Methods

### EMM Dataset and Training

EMM consisted of 5 independently trained 3D convolutional neural networks (CNNs), comprising two versions with different numbers of trainable parameters: a large version utilizing ResNet[44] 101 and 152, and DenseNet[45] 121, 169, and 201; and a small version employing ResNet 18, 34, 50, 101, and 152. These networks were initialized using 2D ImageNet[46] pre-trained weights and adapted to 3D via the Inflated 3D (I3D)[47] method, which has shown success previously[42]. EMM sub-models were trained using the open-source RSNA 2019 ICH Detection Challenge[34,48] dataset and was evaluated using an independent dataset collected at our institution. We trained the models on different subsets of the RSNA dataset, including 18,370 (100%), 9,185 (50%), 4,592 (25%), and 918 (5%) studies, to evaluate EMM's performance across varying training data sizes. All subsets had an ICH prevalence of about 41%. Each model was trained for 100 epochs to ensure convergence with the Adam optimizer and a learning rate of $10^{-4}$. All training was conducted on a server of four NVIDIA L40 GPUs using the PyTorch Lightning framework. Based on GPU memory constraints (48 GB), we set the batch size to 4 for ResNet models and 2 for DenseNet models.

### EMM Evaluation Dataset

We evaluated the EMM using a dataset of 2,919 CT studies (1,315 ICH-positive and 1,604 ICH-negative, 45% ICH prevalence) with a balanced gender distribution (50.1% male, 49.8% female) and a wide age range (0.16-104.58 years; median: 67.13 years; interquartile range: 49.65-80.00 years). Since AI model performance is known to vary with disease prevalence[49,50], we evaluated both the primary AI and EMM performance across different prevalence levels. A recent internal evaluation at our institution covering 8,935 studies between July and November 2024 revealed ICH prevalences of 34.77% for in-patient, 9.09% for out-patient, and 6.52% for emergency units, with an overall average prevalence of 16.70%. Based on these observations, we selected three representative prevalence levels for evaluation: 30%, 15%, and 5%.

### EMM Training

To prepare input data for the EMM, we preprocessed all non-contrast axial head CT DICOM images using the Medical Open Network for AI (MONAI) toolkit[51]. The preprocessing pipeline consisted of several standardization steps: reorienting images to the "left-posterior-superior" (LPS) coordinate system, normalizing the in-plane resolution to 0.45mm, and resizing (either cropping or padding depending on the matrix size) the in-plane matrix dimensions to 512×512 pixels using PyTorch's adaptive average pool method, while preserving the original slice resolution. During training, we employed random cropping in the slice dimension, selecting a contiguous block of 30 slices. For testing, we used a sliding window of 30 slices and averaged the ICH SoftMax probabilities across overlapping windows to generate the final prediction.

*ICH Detection AI models*

We evaluated EMM's monitoring capabilities on two distinct ICH detection AI systems: an FDA-cleared commercial product and an open-source model that secured second place in the RSNA 2019 ICH detection challenge[34,48].

The FDA-cleared model is a black-box system with undisclosed training data and architecture that provides binary labels for presence of ICH and identifies suspicious slices. We monitored this model using EMM trained on the complete (100%, N=18,370) RSNA 2019 ICH Detection Challenge dataset. While this dataset's license restricts usage to academic and non-commercial purposes, we do not have access to information regarding whether the FDA-cleared ICH AI model utilized this dataset during its development.

The open-source RSNA 2019 Challenge second-place model[33] employs a 2D ResNext-101[52] network for slice-level feature extraction, followed by two levels of Bidirectional LSTM networks for feature summarization and ICH detection. We selected the second-place model rather than the first-place winner because retraining the top model would require extensive time while offering only marginal performance improvement (≤2.3%) based on the leaderboard. Although the original open-source model was trained on the complete RSNA Challenge dataset, we retrained it using only 50% of the data and reserved the remaining 50% for EMM training. This simulates real-world deployment scenarios where the primary ICH detector and EMM are trained on different datasets. For both the FDA-cleared and the open-source ICH detection models, no additional image preprocessing was performed and the original DICOM was sent as the input.

*Manually Annotated Sub-dataset for Shapley Analysis*

To comprehensively analyze features that drive high EMM agreement, we manually annotated a smaller dataset (N=281), including ICH segmentation, volume measurements, and identification of mimicking imaging features. This curated dataset comprised 210 ICH-positive and 71 ICH-negative subjects and their associated studies. The ICH-positive cases span 7 distinct ICH subtypes: subdural (SDH, N=35), subarachnoid (SAH, N=50), epidural (EDH, N=15), intraparenchymal (IPH, N=19), intraventricular (IVH, N=2), diffuse axonal injury (DAI, N=1), and multi-compartmental hemorrhages (Multi-H, N=88). Among the 71 ICH-negative cases, 43 cases were specifically selected to include features that mimic hemorrhages (e.g., hyper-density such as calcification or tumor), while 28 were from normal subjects. A neuroradiology fellow reviewed and validated all clinical labels to ensure accurate ground truth for our analysis.

*Comprehensive List of Features for Shapley Analysis*

In Shapley analysis, we prepared a comprehensive list of features including pathology-related metrics (ICH volume and type), patient characteristics (brain volume, age, gender), positioning parameters (rotation, translation), image acquisition parameters (pixel spacing, slice thickness, kVp, X-ray tube current, CT scanner manufacturer), and image reconstruction parameters (reconstruction convolution kernel and filter type).

### *Explaining Features Contributing to High EMM Agreement using Shapley Analysis*

To elucidate the features contributing to the high level of agreement between EMM sub-models and the primary AI model, we conducted Shapley analysis[35] using the Python "shap" package (v0.46.0). This analysis employed an XGBoost[53] (v2.1.1) classifier to learn the relationship between feature values and EMM agreement and to evaluate the importance of each feature leading to high EMM agreement, quantified by the probability ranges between 0 to 1. Higher Shapley values indicate features important for 100% EMM agreement.

### *ICH Volume Estimation for Shapley Analysis*

To evaluate whether ICH volume influences EMM monitoring performance, we implemented a systematic protocol for ICH volume estimation. First, we employed Viola-UNet[54], the winning model from the Instance 2022 ICH Segmentation Challenge[55,56], to generate initial ICH segmentations. A radiology resident reviewed these segmentations and marked any errors directly on the images. A trained researcher then manually corrected the marked discrepancies using 3D Slicer software (version 5.6.2) to ensure accurate hemorrhage delineation. Finally, we calculated ICH volumes using the corrected ICH masks and image resolution data from the DICOM headers.

### *Estimating Patient Brain Volume and Orientation Information for Shapley Analysis*

Since hemorrhage detection can be challenging in brains of different sizes or certain brain orientations, we analyzed brain volume and orientation as potential factors affecting EMM performance, alongside the previously mentioned features. Using the FMRIB Software Library[57] (FSL 6.0.7.13), we developed an automated pipeline following an established protocol[58] to extract brain masks and estimate brain volumes. We then employed FSL FLIRT (FMRIB's Linear Image Registration Tool) to perform 9-degree-of-freedom brain registration, aligning each image to the MNI 2019b non-symmetrical T1 brain template. The resulting rotation, translation, and scaling parameters were incorporated into our Shapley analysis as quantitative measures of brain orientation.

### *Analysis of the tradeoff between false alarm rate and the relative accuracy improvement*

When the decreased confidence group in **Figure 1b** is further reviewed by radiologists, some cases may actually be found to be labeled correctly by the primary model; we consider these cases to be false alarms. The false-alarm rate is defined as the percentage of unnecessary reviews of correctly labeled cases. After further reviewing the cases in the decreased confidence group, we assumed that the radiologists will always correctly label the cases, improving overall accuracy. We define relative improvement in accuracy as the percentage increase in accuracy after reviewing the decreased confidence group compared to the baseline accuracy of the primary model.

*Statistical Analysis*

To assess the reliability of our model's performance metrics, we calculated 95% confidence intervals (CIs) using bootstrapping. We conducted 1,000 random draws with replacement from the set of ground-truth labels and corresponding model predictions. To create evaluation dataset at target prevalence levels (30%, 15%, and 5%) different from the original distribution (45%), we down-sampled ICH-positive and resampled ICH-negative cases. For example, to create datasets with a controlled 30% prevalence of ICH-positive cases, we performed random sampling with replacement from our original dataset. Specifically, we randomly selected $0.3 \times N_n$ ICH-positive cases and $N_n$ ICH-negative cases (where $N_n$ represents the total number of ICH-negative cases in the original dataset). After each draw, we computed key performance metrics such as sensitivity, positive predictive value (PPV), specificity, and negative predictive value (NPV). We then determined the 95% confidence intervals by identifying the 2.5th and 97.5th percentiles of these metrics across all bootstrap iterations. To test the significance of differences in metric between different groups, bootstrapping was also applied to estimate the p-value, with the null hypothesis that there is no difference between the two paired groups.

# Supplementary Information

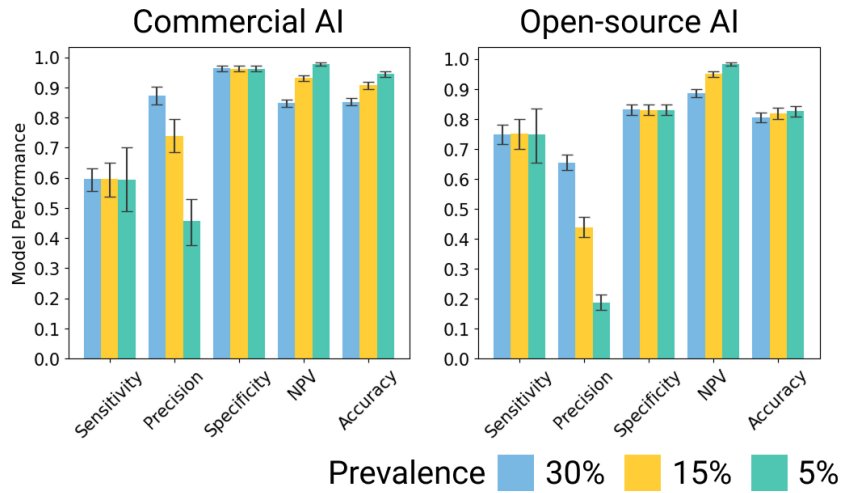

***Supplementary Figure 1. Primary ICH detection AI performance at different ICH prevalences.***
*Two primary ICH detection models were evaluated in this study: an FDA-cleared model and an open-source model. The FDA-cleared model demonstrated relatively lower sensitivity and NPV, but higher precision, specificity, and overall accuracy compared to the open-source model. As prevalence decreased from 30% to 5%, both models exhibited decreased precision with increased NPV and accuracy, while sensitivity and specificity remained unchanged.*

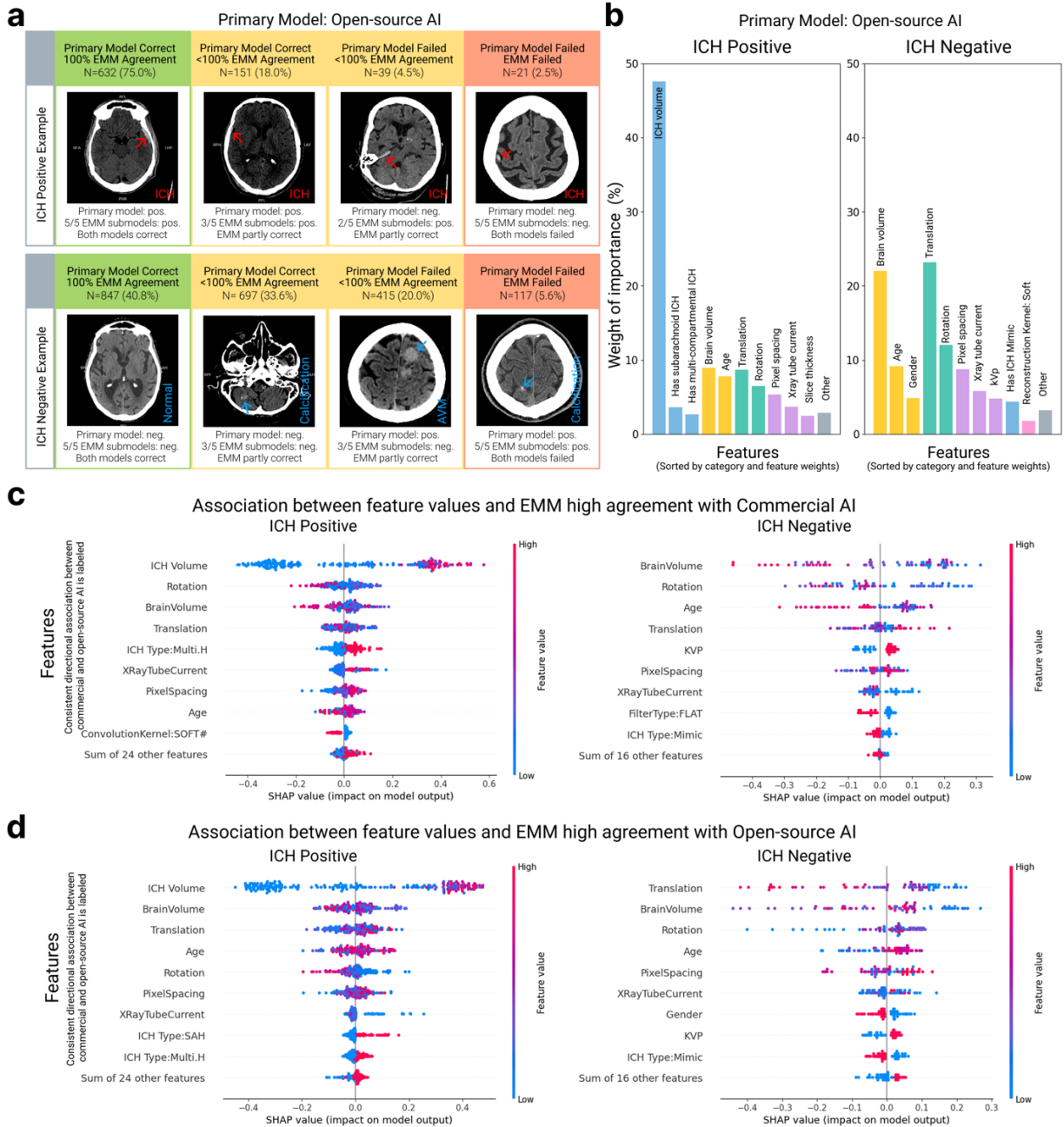

***Supplementary Figure 2. a.*** *Example cases for which EMM showed different levels of agreement with the Open-Source primary ICH detection model.* ***b.*** *Quantitative analysis on the importance of features affecting EMM agreement with the commercial primary model in ICH-positive and ICH-negative cases. c. Shapley analysis on feature values' association with 100% EMM agreements for the commercial AI. d. Shapley analysis on feature values' association with 100% EMM agreements for the open-source AI.*

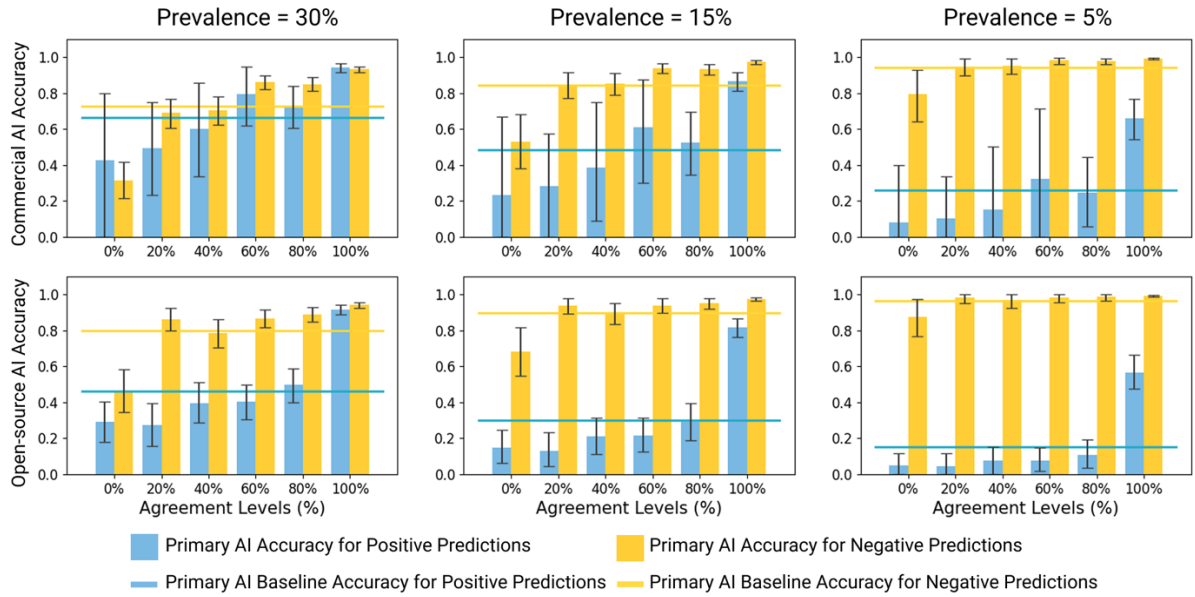

*Supplementary Figure 3. EMM level of agreement stratified primary AI predictions into categories of different accuracies.* The primary model's accuracy for both positive (blue) and negative (yellow) predictions is shown across different EMM agreement and prevalence levels.

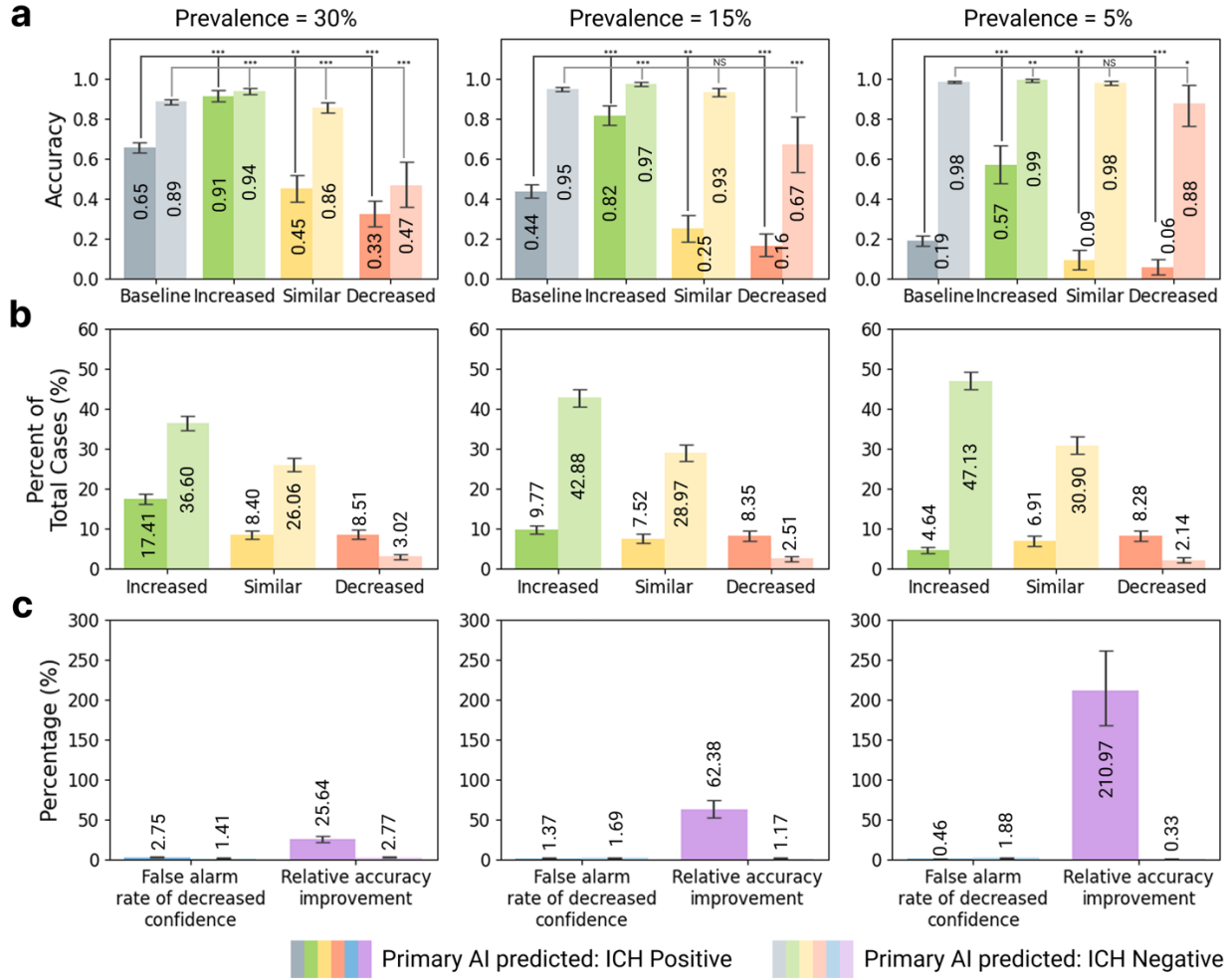

***Supplementary Figure 4. EMM stratified cases into increased, similar, and decreased confidence groups for open-source primary AI and enables customized clinical decision making.** **a.** Accuracy of each subgroups. **b.** Percent of cases of each subgroups. **c.** Tradeoff between the unnecessary false alarm review and the relative accuracy improvement. Same EMM agreement thresholds were used as in **Figure 1b**.*

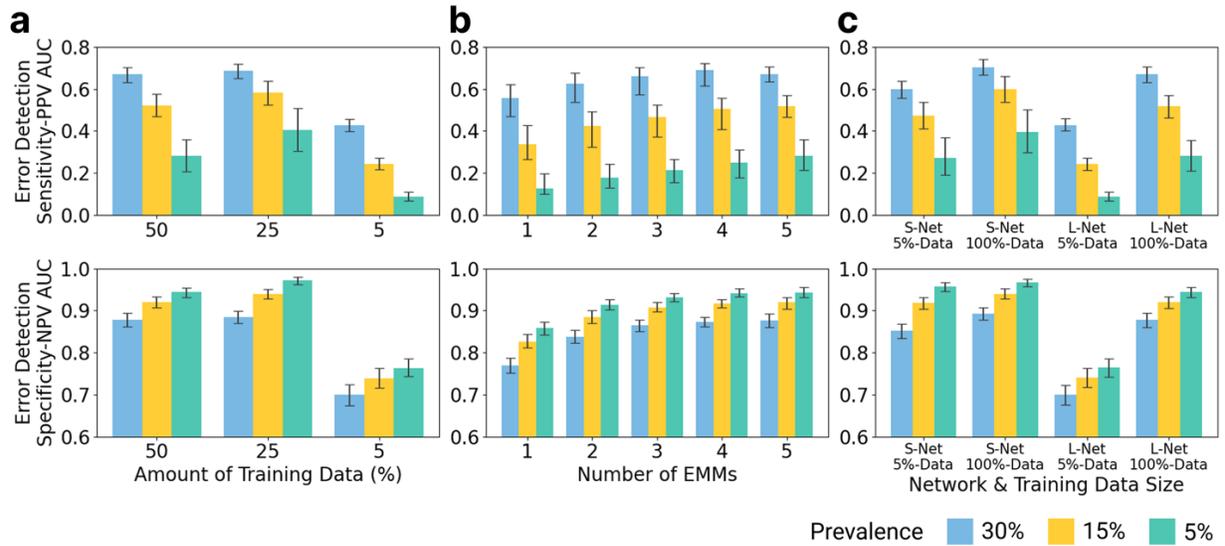

*Supplementary Figure 5. EMM performance on catching open-source ICH AI errors generally improves with increasing training data volume (a), number of sub-models (b), and sub-model sizes.*

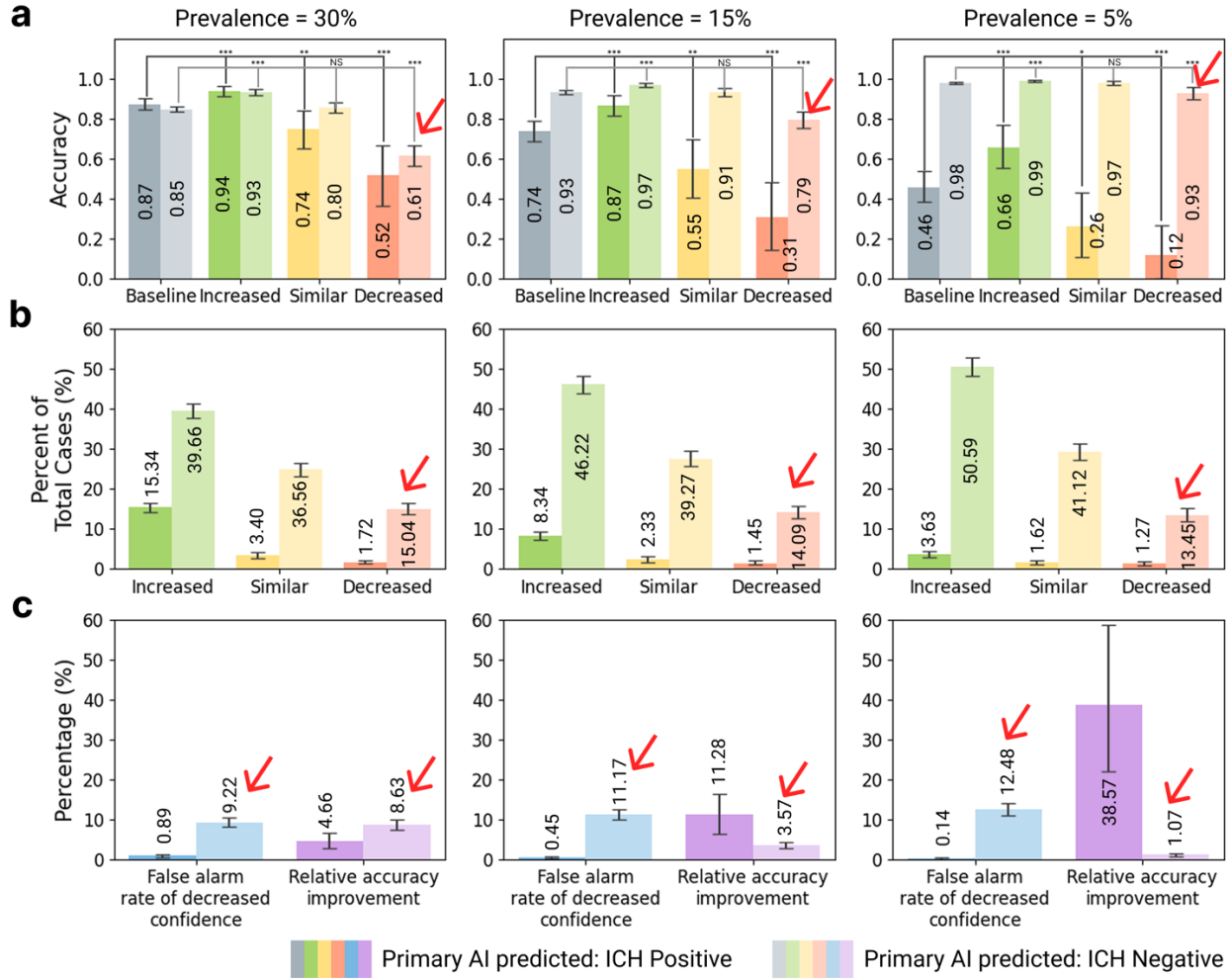

***Supplementary Figure 6. Suboptimal EMM stratification threshold increased unnecessary reviews for false alarm cases while reduced the relative accuracy gain.*** *Red arrows indicate the changes compared to Figure 3. In this example, the negative yellow thresholds changed from [80%, 60%, 40%, and 20%] to [80% and 60%], and negative red thresholds changed from [0%] to [40%, 20%, and 0%] compared to **Figure 3a**.*